Published as a paper at the 2nd DeLTa Workshop, ICLR 2026

# Paired Wasserstein Autoencoders for Conditional Sampling

**Moritz Piening**
Institute of Mathematics
Technische Universität Berlin
Straße des 17. Juni 136, D-10623 Berlin, Germany
piening@math.tu-berlin.de

**Matthias Chung**
Department of Mathematics
Emory University
400 Dowman Drive, Atlanta, GA 30322, USA
matthias.chung@emory.edu

## Abstract

Generative autoencoders learn compact latent representations of data distributions through jointly optimized encoder–decoder pairs. In particular, Wasserstein autoencoders (WAEs) minimize a relaxed optimal transport (OT) objective, where similarity between distributions is measured through a cost-minimizing joint distribution (OT coupling). Beyond distribution matching, neural OT methods aim to learn mappings between two data distributions induced by an OT coupling. Building on the formulation of the WAE loss, we derive a novel loss that enables sampling from OT-type couplings via two paired WAEs with shared latent space. The resulting fully parametrized joint distribution yields (i) learned cost-optimal transport maps between the two data distributions via deterministic encoders. Under cost-consistency constraints, it further enables (ii) conditional sampling from an OT-type coupling through stochastic decoders. As a proof of concept, we use synthetic data with known and visualizable marginal and conditional distributions.

## 1 Introduction

Generative autoencoders learn low-dimensional representations of data distributions via paired encoder–decoder architectures (Kingma et al., 2019). Wasserstein autoencoders (WAEs) refine this approach by replacing divergence-based matching with a relaxed optimal transport (OT) objective based on cost-minimizing couplings (Tolstikhin et al., 2018; Xu et al., 2019; Kim et al., 2022).

Beyond distribution approximation, optimal transport offers a principled way to relate two data distributions through learned couplings and transport maps. Recent neural OT methods focus on recovering such cost-optimal mappings, including (Korotin et al., 2023; Tong et al., 2023; Kim et al., 2024; De Bortoli et al., 2024; Chemseddine et al., 2025; Wald & Steidl, 2025). However, they typically neither parametrize the joint distribution nor enable sampling from the coupling.

In order to obtain a compact representation, we employ the idea of coupled autoencoders (Zeng et al., 2015; Hao & Shafto, 2023; Hart et al., 2025; Chung et al., 2025). Thus, we introduce paired Wasserstein autoencoders, which jointly model two distributions using coupled WAEs with a shared latent space. Inspired by the WAE loss derivation, we derive an analogous objective for the OT coupling, leading to an unbalanced OT-type loss over the joint distribution. This construction yields a fully parametrized approximation of OT-type couplings, enabling both learned cost-optimal transport maps via deterministic encoders and conditional sampling via stochastic encoders. We validate the approach on synthetic data with known and visualizable marginal and conditional distributions.

## 2 Optimal Transport and Wasserstein Autoencoders

We introduce OT background of WAEs using Markov kernels (Villani et al., 2008). Throughout, let $X, Y, Z, Z'$ denote complete separable metric spaces, such as $\mathbb{R}^d$. We denote by $\mathcal{P}(X)$ the set of Borel probability measures on $X$ and by $\mathcal{P}_2(X) \subset \mathcal{P}(X)$ the subset with finite second moments.

**Markov Kernels and Optimal Transport.** Given $\mu \in \mathcal{P}_2(X)$ and $\nu \in \mathcal{P}_2(Y)$, a Markov kernel from $X$ to $Y$ is a measurable map $K : X \to \mathcal{P}(Y)$, such that for every Borel set $A \in \mathcal{B}(Y)$ the map $x \mapsto K(x)(A)$ is measurable. For a measurable map $T : X \to Y$ and a measure $\mu$ on $X$, the push-forward $T_\# \mu$ is defined by $(T_\# \mu)(A) := \mu(T^{-1}(A))$ for $A \in \mathcal{B}(Y)$. This definition extends to Markov kernels. Given a Markov kernel $K : X \to \mathcal{P}(Y)$ and a measure $\mu$ on $X$, the push-forward $K_{\#}\mu$ is the measure on $Y$ defined by $(K_{\#}\mu)(A) \coloneqq \int_X K(x)(A) \, \mathrm{d}\mu(x)$ for $A \in \mathcal{B}(Y)$.

A transport plan (or coupling) between $\mu \in \mathcal{P}_2(X)$ and $\nu \in \mathcal{P}_2(Y)$ is a measure $\pi \in \Pi(\mu, \nu) \subset \mathcal{P}_2(X \times Y)$ satisfying $(\mathrm{proj}_X)_{\#}\pi = \mu$ and $(\mathrm{proj}_Y)_{\#}\pi = \nu$, where $\mathrm{proj}_X(x, y) = x$ and $\mathrm{proj}_Y(x, y) = y$ denote the canonical projections. Any transport plan $\pi \in \mathcal{P}_2(X \times Y)$ with first marginal $\mu$ admits a disintegration with respect to $\mu$, that is, there exists a $\mu$-almost everywhere uniquely defined Markov kernel $K : X \to \mathcal{P}(Y)$ such that

$$\int_{X \times Y} f(x, y) \, \mathrm{d}\pi(x, y) = \int_X \int_Y f(x, y) \, \mathrm{d}K(x)(y) \, \mathrm{d}\mu(x) \quad \text{for all } \pi\text{-integrable } f \in L^1(\pi).$$

Conversely, under moment conditions, a Markov kernel $K : X \to \mathcal{P}(Y)$ defines a measure $\pi \in \mathcal{P}_2(X \times Y)$ by the above identity, whose marginals satisfy $(\mathrm{proj}_X)_{\#}\pi = \mu$ and $(\mathrm{proj}_Y)_{\#}\pi = K_{\#}\mu$. In particular, transport plans between $\mu$ and $\nu$ are in one-to-one correspondence (within $\mu$-almost everywhere equivalence) with Markov kernels $K$ satisfying $K_{\#}\mu = \nu$.

Given a measurable, lower-semicontinuous, and lower-bounded cost function $c : X \times Y \to \mathbb{R}$, we can now define the Kantorovich OT problem between $\mu \in \mathcal{P}_2(X)$ and $\nu \in \mathcal{P}_2(Y)$ given by

$$\mathrm{OT}_c(\mu, \nu) \coloneqq \inf_{\pi \in \Pi(\mu, \nu)} \int_{X \times Y} c(x, y) \, \mathrm{d}\pi(x, y) = \inf_{\substack{K : X \to \mathcal{P}(Y) \\ K_{\#}\mu = \nu}} \int_X \int_Y c(x, y) \, \mathrm{d}K(x)(y) \, \mathrm{d}\mu(x). \quad (1)$$

For $c_p(x, y) = \|x - y\|^p$ and $X = Y = \mathbb{R}^d$, this defines the Euclidean $p$-Wasserstein distance via $\mathrm{W}_p^p(\mu, \nu) \coloneqq \mathrm{OT}_{c_p}(\mu, \nu)$. If an optimal kernel satisfies $K^*(x) = \delta_{M^*(x)}$ for $\mu$-almost every $x$, then $M^* : X \to Y$ is an optimal Monge transport map.

**Wasserstein decoders.** Latent spaces naturally fit into this formulation. Let $\mu \in \mathcal{P}_2(X)$ be a data distribution and $\xi \in \mathcal{P}_2(Z)$ a fixed latent prior. A (deterministic) Wasserstein decoder is a measurable map $D : Z \to X$, inducing the push-forward $D_{\#}\xi$. While direct optimization of $D$ is difficult, an optimal Wasserstein decoder would minimize $\mathrm{OT}_c(\mu, D_{\#}\xi)$ over all such maps $D$.

However, given a fixed $D$, we can reparameterize the OT cost based on $\xi$ (Tolstikhin et al., 2018). In particular, the Kantorovich objective becomes

$$\mathrm{OT}_c(\mu, D_{\#}\xi) = \inf_{\substack{K : X \to \mathcal{P}(Z) \\ K_{\#}\mu = \xi}} \int_X \int_Z c(x, D(z)) \, \mathrm{d}K(x)(z) \, \mathrm{d}\mu(x). \quad (2)$$

This kernel formulation naturally induces an autoencoder by interpreting the Markov kernel as a stochastic encoder. In addition to a parametrized decoder $D_\theta$, let $E_\theta : X \times Z' \to Z$ be a parametrized measurable map and let $\xi' \in \mathcal{P}(Z')$ be a fixed auxiliary probability measure. The associated Markov kernel $K_\theta : X \to \mathcal{P}(Z)$ is then defined by

$$K_\theta(x) \coloneqq (E_\theta(x, \cdot))_{\#}\xi', \qquad x \in X.$$

In practice, Wasserstein autoencoders are now trained by minimizing the relaxed OT problem

$$\min_\theta \underbrace{\int_X \int_{Z'} c\big(x, D_\theta(E_\theta(x, z'))\big) \, \mathrm{d}\xi'(z') \, \mathrm{d}\mu(x) \; + \; \lambda_{\mathrm{WAE}} \, \widehat{\mathrm{StatDiv}}\big((E_\theta)_{\#}(\mu \otimes \xi') \,\|\, \xi\big)}_{=: \widehat{\mathcal{T}}_c(\mu, E_\theta, D_\theta)}, \quad (3)$$

where the divergence term enforces the aggregated latent constraint $(E_\theta)_{\#}(\mu \otimes \xi') = (K_\theta)_{\#}\mu \approx \xi$. In particular, we employ a computable minibatch estimator for some true statistical divergence $\mathrm{StatDiv}(E_{\theta,\#}(\mu \otimes \xi') \,\|\, \xi)$. In practice, this divergence is set to be an OT-type estimator, a maximum mean discrepancy or an adversarial network (Tolstikhin et al., 2018; Kolouri et al., 2019; Xu et al., 2019). Lastly, note that setting $E_\theta(x, z') = E_\theta(x)$ everywhere results in a deterministic encoder.

## 3 Paired Wasserstein Autoencoders

Conceptually, WAEs formulate the search for an OT-optimal generator for $\mu$ as an optimization problem over stochastic encoders $K : X \to \mathcal{P}(Z)$ by relaxing $K_{\#}\mu = \xi$. Here, present how we can apply the same idea to estimate the transport between $\mu$ and $\nu$ based on shared latent reference measure $\xi$ and decoders $D_\mu : Z \to X$ and $D_\nu : Z \to Y$. Within this paired autoencoder framework, transport between $\mu$ and $\nu$ is approximated by coupling both distributions through the common latent space $Z$. Note that resulting objectives introduced below are surrogate objectives, merely serve to derive OT approximations rather than exact OT solutions and solutions might not exist.

**Paired WAEs.** Our goal is a parametric approximation of a cost-optimal joint distribution between $\mu \in \mathcal{P}_2(X)$ and $\nu \in \mathcal{P}_2(Y)$, i.e., we seek a transport plan $\pi \in \Pi(\mu,\nu)$ minimizing $\int_{X\times Y} c(x,y)\,\mathrm{d}\pi(x,y)$. Analogously to (2), we consider $\xi \in \mathcal{P}_2(Z)$ and a pair of measurable decoders $(D_\mu, D_\nu) : Z \to X \times Y$. Such a decoder pair induces a joint distribution $(D_\mu, D_\nu)_\#\xi \in \mathcal{P}_2(X \times Y)$, i.e., a generative parametrization of a transport plan. Consequently, our OT-type objective is to find decoders $(D_\mu, D_\nu)$ that realize

$$\inf_{D_\mu, D_\nu} \int_Z c(D_\mu(z), D_\nu(z))\,\mathrm{d}\xi(z), \qquad \text{such that} \quad D_{\mu,\#}\xi = \mu, D_{\nu,\#}\xi = \nu.$$

While a solution might not exist, we can again relax the marginal conditions to arrive at

$$\inf_{D_\mu, D_\nu} \int_Z c(D_\mu(z), D_\nu(z))\,\mathrm{d}\xi(z) + \lambda_{\text{joint}}\,\mathrm{W}_p^p(D_{\mu,\#}\xi, \mu) + \lambda_{\text{joint}}\,\mathrm{W}_p^p(D_{\nu,\#}\xi, \nu).$$

Again, replacing the Wasserstein distances $\mathrm{W}_p^p(\mu,\nu) = \mathrm{OT}_{c_p}(\mu,\nu)$ with the parametric WAE problem (3), we get our double-relaxed *paired WAE* problem

$$\min_\theta \underbrace{\int c\big(D_{\mu,\theta}(z), D_{\nu,\theta}(z)\big)\ \mathrm{d}\xi(z) + \lambda_{\text{joint}}\widehat{\mathcal{T}}_{c_p}\big(\mu, E_{\mu,\theta}, D_{\mu,\theta}\big) + \lambda_{\text{joint}}\widehat{\mathcal{T}}_{c_p}\big(\nu, E_{\nu,\theta}, D_{\nu,\theta}\big)}_{=:\mathcal{L}(\theta)}. \tag{4}$$

The objective (4) can be interpreted as an unbalanced OT formulation that couples $\mu$ and $\nu$ through a shared latent variable while softly enforcing marginal consistency via WAE regularization. Unlike a classical WAE, which approximates a single target distribution through a 'semi-unbalanced' OT problem, the paired Wasserstein autoencoder provides a parametric approximation of an unbalanced OT coupling between $\mu$ and $\nu$. The resulting loss is approximated via samples from $\mu$, $\nu$, $\xi$, and $\xi'$.

**Neural Monge Mapping via Deterministic Encoders.** Our paired decoders induce a joint distribution $(D_{\mu,\theta}, D_{\nu,\theta})_\#\xi \in \mathcal{P}_2(X \times Y)$ that approximates an optimal plan realizing (1). Under assumption of existence and invertibility ($D_{\mu,\theta}(E_{\mu,\theta}(x)) = x$, $D_{\nu,\theta}(E_{\nu,\theta}(y)) = y$) for **deterministic** encoders, we would further obtain deterministic cost-minimizing Monge-like maps

$$x \mapsto M_\theta^{X\to Y}(x) := D_{\nu,\theta}(E_{\mu,\theta}(x)), \qquad y \mapsto M_\theta^{Y\to X}(y) := D_{\mu,\theta}(E_{\nu,\theta}(y)).$$

Under existence assumption of a Monge map, we can approximate it for general costs, see Sec. 4.1.

**Conditional Sampling for Inverse Problems** Given a linear inverse problem $y = f(x) + \varepsilon$ with a linear map $f : \mathbb{R}^{d_X} \to \mathbb{R}^{d_Y}$, spherical Gaussian noise $\varepsilon$, marginals $x \sim \mu$, $y \sim \nu$ and joint $(x,y) \sim \pi$, we are interested in the conditional measure $\mu_y$, i.e., the probability measure on $X$ obtained by conditioning the $\pi$ on the observation $Y = y$, see (Stuart, 2010; Arridge et al., 2019; Piening et al., 2024). While $\pi$ is not a OT plan, a **heuristic** should minimize $\mathrm{OT}_{c_f}(\mu,\nu)$ for

$$c_f(x,y) = c_2(f(x), y) = \|f(x) - y\|^2$$

in some form. Minimizing (4) for $c_f$ and stochastic encoders $\xi' \in \mathcal{P}_2(\mathbb{R}^{d_X})$, we obtain conditional-like distributions induced by the learned coupling $(D_{\mu,\theta} \circ E_{\nu,\theta}(y,\cdot))_\#\xi'$, see Sec. 4.2.

## 4 Experiments

We use random Gaussian mixtures (GMMs) in $\mathbb{R}^d$ by uniform sampling of 8 means in $[-1,1]^d$, diagonal covariances with variances in $[0.1, 0.2]$ and normalized weights. Both encoders and decoders are 2-hidden-layer MLPs (enc.: $d \to 256 \to 128 \to 2$, dec.: $2 \to 128 \to 256 \to d$, Adam lr: 0.001). We use 50,000 samples per marginal. Training uses batch size 256 per $\mu$, $\nu$ and $\xi$ for 50 epochs. $\mathrm{StatDiv}$ is the empirical energy distance (Székely & Rizzo, 2013), see (4.1). We average results for each experiment over 5 random seeds. Each is evaluated with 100 random batches of size 512. We generally employ $\hat{\mathcal{T}}_{c_2}$ in (4) for both experiments based on an Euclidean data representation.

### 4.1 Neural OT Map on Sphere

**Setup.** We approximate OT between two distributions on the sphere $\mathbb{S}^2$, where we sample two GMMs in $\mathbb{R}^3$ and $\ell_2$-normalize to $\mathbb{S}^2$. We enforce spherical decoder outputs by $\ell_2$-normalizing and employ $\xi \sim \mathcal{U}([0,1]^2)$. We train on each mixture pair for different combinations of $\lambda_{\text{joint}}$ and $\lambda_{\text{WAE}}$.

| Parameters | | Spherical OT Quality | | |
|---|---|---|---|---|
| $\lambda_{\text{joint}}$ | $\lambda_{\text{WAE}}$ | MG ↓ | ED ↓ | MG + 10 · ED ↓ |
| 0 | 0 | $0.00 \pm 0.0$ | $1.64 \pm 0.2$ | $16.45 \pm 1.7$ |
| 10 | 0 | $0.07 \pm 0.0$ | $0.87 \pm 0.2$ | $8.73 \pm 1.6$ |
| 10 | 10 | $0.18 \pm 0.2$ | $0.02 \pm 0.0$ | **0.39**$\pm$ 0.2 |

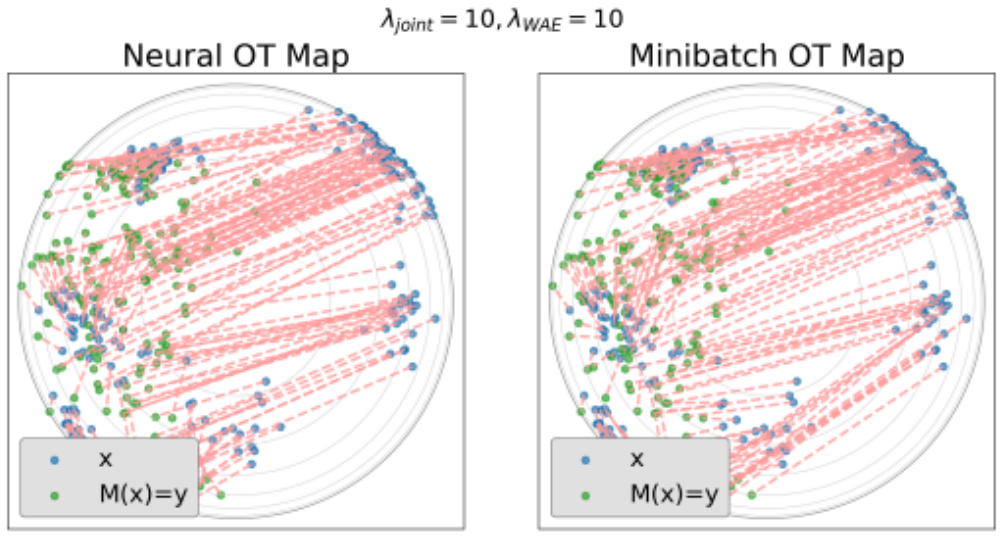

Table 1: **Left**: Effect of $\lambda_{\text{joint}}$ and $\lambda_{\text{WAE}}$ on mapping cost via Monge gap (MG) and generative quality via energy distance (ED). **Right**: Visualization of our neural OT map (left) between spherical $\mu$ and $(D_{\nu,\theta} \circ E_{\mu,\theta})_{\#}\mu$ and the minibatch OT map (right). Both OT mappings (orange) are very similar.

**Evaluation.** Given minibatches $(x_i)_{i=1}^n \sim \mu$, $(y_i)_{i=1}^n \sim \nu$, $(z_k)_{k=1}^n \sim \xi$, we define the directional Monge gap (Uscidda & Cuturi, 2023) of $M_\theta^{X\to Y}$ for $X \to Y$ as

$$\mathrm{MG}^{X\to Y}(x, M_\theta^{X\to Y}) = \frac{1}{n}\sum_{i=1}^{n} c(x_i, M_\theta^{X\to Y}(x_i)) - \min_{\sigma\in S_n} \frac{1}{n}\sum_{i=1}^{n} c(x_i, M_\theta^{X\to Y}(x_{\sigma(i)})),$$

where $\sigma \in S_n$ is the cost-miniming permutation estimated using POT (Flamary et al., 2021). We set the final Monge Gap as the average of the $X \to Y$ and the $Y \to X$ Monge gap. Evaluating generative fit, the (batched) Energy distance (ED) (Székely & Rizzo, 2013; Hertrich et al., 2024) is

$$\widehat{\mathrm{ED}}_\mu(x, \tilde{x}) = \frac{2}{n^2}\sum_{i,j=1}^{n} c_1(x_i, \tilde{x}_j) - \frac{1}{n^2}\sum_{i,j=1}^{n} c_1(x_i, x_j) - \frac{1}{n^2}\sum_{i,j=1}^{n} c_1(\tilde{x}_i, \tilde{x}_j), \qquad \tilde{x}_i := D_{\mu,\theta}(z_i).$$

Our final energy is again the mean energy for $\mu$ and $\nu$. Note that while our $x_i, y_j \in \mathbb{S}^2$ originally, we here use the canonical embedding to $\mathbb{R}^3$ for calculation of the ED.

**Results.** Table 1 evaluates and visualizes the learned transport mappings. While small deviations from the exact maps are visible, the overall transport structure is preserved. Both $\lambda_{\text{joint}}$ and $\lambda_{\text{WAE}}$ yield large MG values, primarily due to violations of the marginal OT constraints, as reflected by the high energy distance. In particular, setting $\lambda_{\text{joint}} = \lambda_{\text{WAE}} = 0$ leads to collapse to a constant embedding. In contrast, optimizing the joint objective $\mathrm{MG} + \kappa \cdot \mathrm{ED}$ enforces marginal preservation and cost-optimal transport in the limit $\kappa \to \infty$. Already for $\kappa = 10$, the learned mappings are closest to a Monge map when $\lambda_{\text{joint}} = \lambda_{\text{WAE}} = 10$.

| Model | Unconditional ED ↓ | Conditional ED ↓ |
|---|---|---|
| Paired WAE | $0.01 \pm 0.0$ | $0.14 \pm 0.2$ |
| cVAE | $0.00 \pm 0.0$ | $0.02 \pm 0.0$ |

Table 2: Generation for Inverse Problem.

### 4.2 Posterior Estimation

**Setup.** We simulate a 2D GMM $\mu$ and a linear operator $F \in \mathbb{R}^{1\times 2}$ drawn from normalized Gaussian noise, generating observations $Y = FX + \varepsilon$ with $\varepsilon \sim \mathcal{N}(0, \sigma^2)$, $\sigma = 0.1$. We train a paired WAE ($\lambda_{\text{WAE}} = \lambda_{\text{joint}} = 10$) and a conditional VAE (cVAE; Sohn et al. (2015)) with shared prior $\xi \sim \mathcal{N}(0, I_2)$ and encoder noise $\xi' \sim \mathcal{N}(0, I_2)$. The cVAE KL term is weighted by the known likelihood variance, scaling the MSE reconstruction by $1/(2\sigma^2)$, and we use the cost $c_F(x, y) = \|Fx - y\|^2$. Since $p(x \mid y)$ is available in closed form as a GMM, we directly compare learned posteriors to the analytic solution, see (Grana et al., 2017; Piening et al., 2024).

**Evaluation.** We report the unconditional Energy Distance (ED) for both marginals using (4.1), and average them to obtain a single score. We additionally report the conditional ED by reporting the mean ED over the sampled learned posterior $D_{\mu,\theta}(E_{\nu,\theta}(y, z))$ and the sampled analytic posterior for 100 different $y$ values.

**Results.** Table 2 shows similar unconditional performance; cVAE is more accurate conditionally but collapses to a 1D conditional (suppl. Table 4), while WAE shows larger support with reasonable conditional quality. Misweighted cVAE KL leads to worse results than our WAE (suppl. Table 3).

## 5 Conclusion

By simplification of hard OT constraints, we introduced a novel WAE-based loss that yields a fully parametrized approximation of an OT-type coupling and validated its behavior in proof-of-concept experiments. While this work mainly serves as a conceptual outline, future work will focus on establishing rigorous formulations, conditions ensuring existence of solutions, mathematical consistency in the limit $\lambda_{\text{WAE}}, \lambda_{\text{joint}} \to \infty$ and extending the approach to real-world applications.

## ACKNOWLEDGMENTS

MP acknowledges the financial support by the German Research Foundation (DFG), GRK2260 BIOQIC project 289347353. This work was partially supported by the National Science Foundation (NSF) under grant DMS-2152661 (MC). We thank Jannis Chemseddine, Paul Hagemann, and Gabriele Steidl (TU Berlin) for fruitful discussion

## SUPPLEMENTARY

### CLOSED-FORM KL WEIGHT FOR SPHERIAL GAUSSIAN

Assume a Gaussian likelihood with fixed variance $\sigma^2$, $p(y \mid x) = \mathcal{N}(y; x, \sigma^2 I)$. Then $\log p(y \mid x) = -\frac{1}{2\sigma^2}\|y - x\|^2 - \frac{d}{2}\log(2\pi\sigma^2)$, and the expected log-likelihood term in the ELBO is $\mathbb{E}_{q(x|y)}[\log p(y \mid x)] = -\frac{1}{2\sigma^2}\mathbb{E}_{q(x|y)}[\|y - x\|^2] - \frac{d}{2}\log(2\pi\sigma^2)$. Ignoring the additive constant, maximizing the ELBO is equivalent to minimizing $\frac{1}{2\sigma^2}\mathbb{E}_{q(x|y)}[\|y - x\|^2] + \mathrm{KL}(q(x \mid y)\|p(x))$.

With a VAE parameterization $q_\theta(x \mid y)$ and $p_\theta(y \mid x) = \mathcal{N}(y; \mu_\theta(x), \sigma^2 I)$, this becomes (up to constants) $\max_\theta \mathcal{L}(\theta; y) \iff \min_\theta \frac{1}{2\sigma^2}\mathbb{E}_{q_\theta(x|y)}[\|y - \mu_\theta(x)\|^2] + \mathrm{KL}(q_\theta(x \mid y)\|p(x))$. In the special case $\mu_\theta(x) = x$, this reduces to $\min_\theta \frac{1}{2\sigma^2}\mathbb{E}_{q_\theta(x|y)}[\|y - x\|^2] + \mathrm{KL}(q_\theta(x \mid y)\|p(x))$.

Introducing a $\beta$-weight on the KL term therefore yields

$$\min_\theta \frac{1}{2\sigma^2}\mathbb{E}_{q_\theta(x|y)}[\|y - \mu_\theta(x)\|^2] + \beta\,\mathrm{KL}(q_\theta(x \mid y)\|p(x)), \tag{5}$$

which deviates from the exact ELBO when $\beta \neq 1$ for fixed-variance Gaussian likelihoods. For further discussion, see Odaibo (2019); Kingma et al. (2019).

### CLOSED-FORM GMM CONDITIONAL/POSTERIOR DISTRIBUTIONS

Based on a GMM prior

$$p_X(x) = \sum_{k=1}^{K} \alpha_k\, \mathcal{N}(x; m_k, \Sigma_k)$$

and a linear Gaussian forward model $Y = FX + \varepsilon$, $\varepsilon \sim \mathcal{N}(0, \sigma^2 I)$, the posterior $p_{X|Y=y}$ is again a Gaussian mixture

$$p_{X|Y=y}(x) = \sum_{k=1}^{K} \tilde{\alpha}_k(y)\, \mathcal{N}(x; \tilde{m}_k, \tilde{\Sigma}_k),$$

| $\beta$ | Unconditional ED | Conditional ED |
|---|---|---|
| Paired WAE | $0.01 \pm 0.0$ | $0.14 \pm 0.2$ |
| 0.01 | $0.02 \pm 0.0$ | $0.10 \pm 0.2$ |
| 0.1 | $0.01 \pm 0.0$ | $0.05 \pm 0.1$ |
| 1 | $0.00 \pm 0.0$ | $0.02 \pm 0.0$ |
| 10 | $0.01 \pm 0.0$ | $0.04 \pm 0.0$ |
| 100 | $0.08 \pm 0.0$ | $0.28 \pm 0.2$ |

Table 3: Ablation of cVAE over $\beta$.

with component-wise parameters

$$\tilde{\Sigma}_k = \left(\Sigma_k^{-1} + \sigma^{-2} F^\top F\right)^{-1}, \qquad \tilde{m}_k = \tilde{\Sigma}_k \left(\Sigma_k^{-1} m_k + \sigma^{-2} F^\top y\right),$$

and posterior mixture weights

$$\tilde{\alpha}_k(y) = \frac{\alpha_k \, \mathcal{N}\big(y;\, F m_k,\, F \Sigma_k F^\top + \sigma^2 I\big)}{\sum_{j=1}^K \alpha_j \, \mathcal{N}\big(y;\, F m_j,\, F \Sigma_j F^\top + \sigma^2 I\big)}.$$

For further details, please see (Grana et al., 2017; Piening et al., 2024)

### Extension for Conditional Sampling Experiment

We display a two estimated conditional (or posterior) distributions in Table 4. We clearly see that the paired WAEs collapse less than the cVAE. We additionally rerun the posterior sampling for different values of $\beta$ in (5) and display the results in Table 3. We see that $\beta = 1$ (our actual choice) is ideal and that large deviation from $\beta = 1$ lead to worse conditional sampling than our WAEs.

### Limitations

This work serves as a conceptual proof of concept that illustrates how transport plans may be approximated via paired neural networks through a relaxation of the OT problem. Therefore, empirical evaluation is intentionally restricted to low-dimensional synthetic data with known and visualizable distributions. In particular, since autoencoder-based generative models such as WAEs and VAEs often face scalability challenges compared to modern flow-based approaches, future work will build on the conceptual foundations established here to develop a competitive flow-based counterpart.

We lastly note that the comparison to cVAEs is not entirely fair, as the paired architecture employs roughly twice the number of parameters. Nevertheless, our model comes with the added benefit of learning a shared joint latent space.

<table>
<tr><th colspan="2">Observation $y_0$</th></tr>
<tr><th>Paired WAE</th><th>cVAE</th></tr>
<tr><td>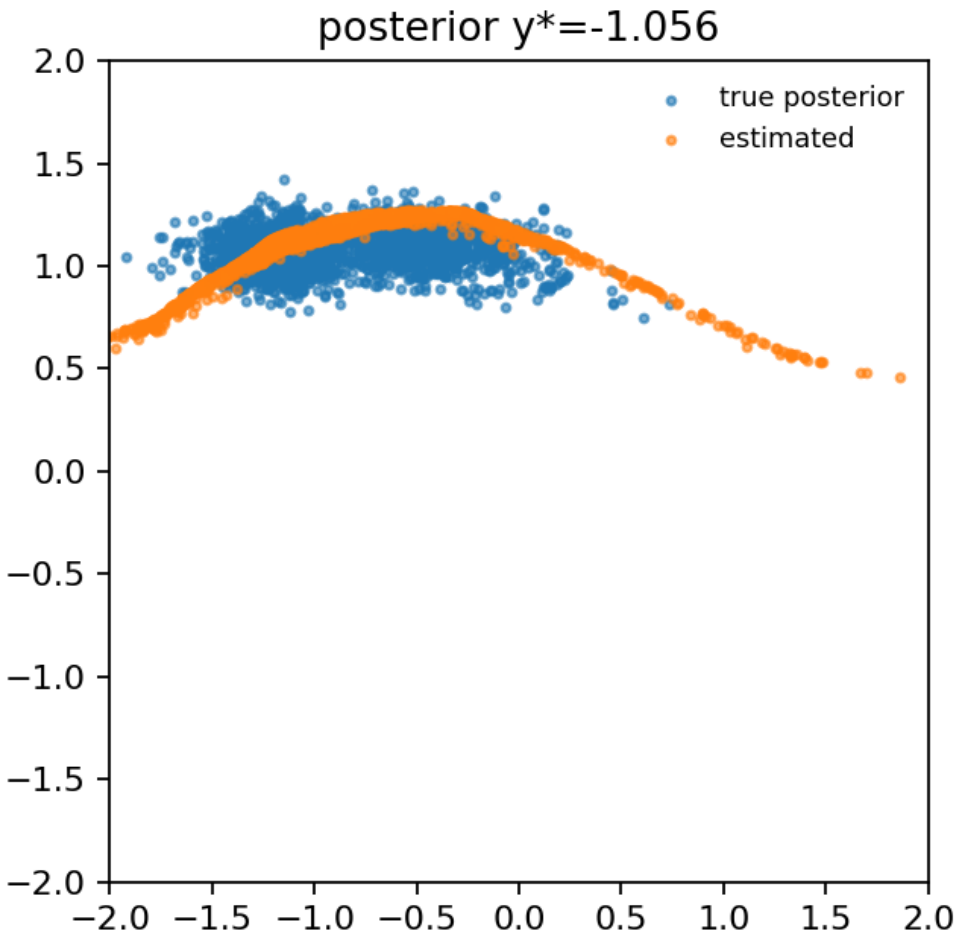
</td><td>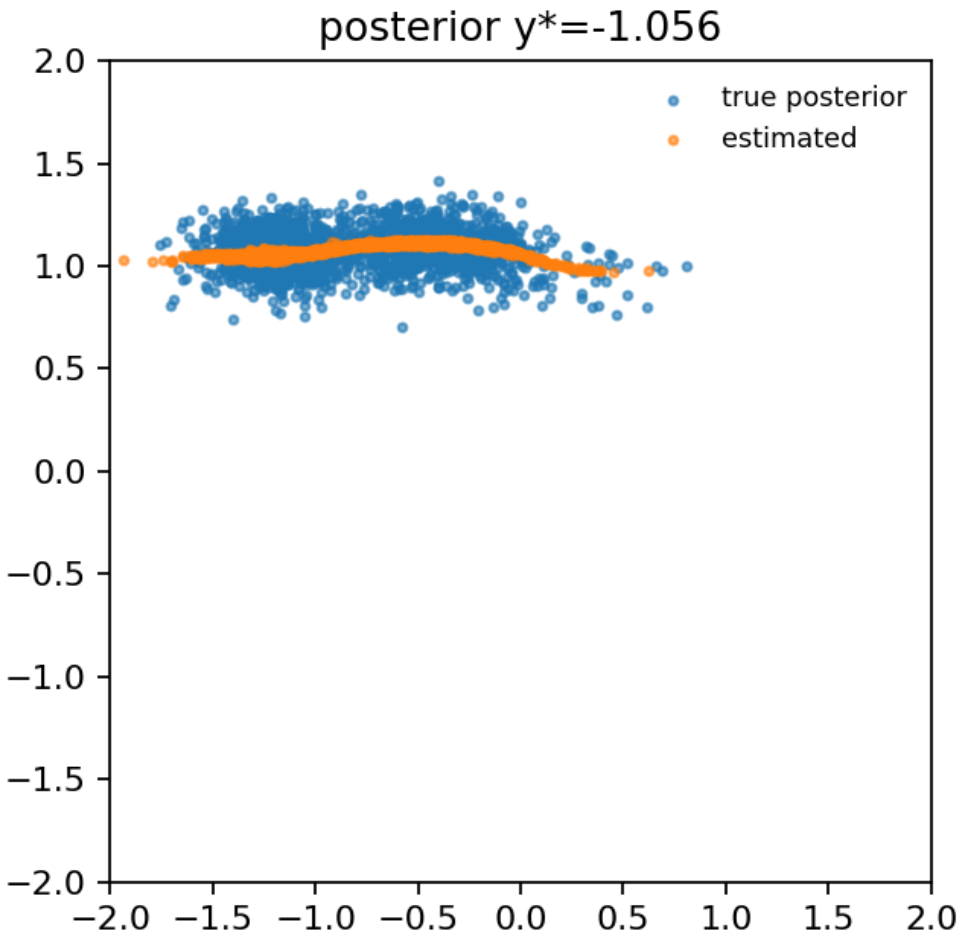
</td></tr>
<tr><th colspan="2">Observation $y_1$</th></tr>
<tr><th>Paired WAE</th><th>cVAE</th></tr>
<tr><td>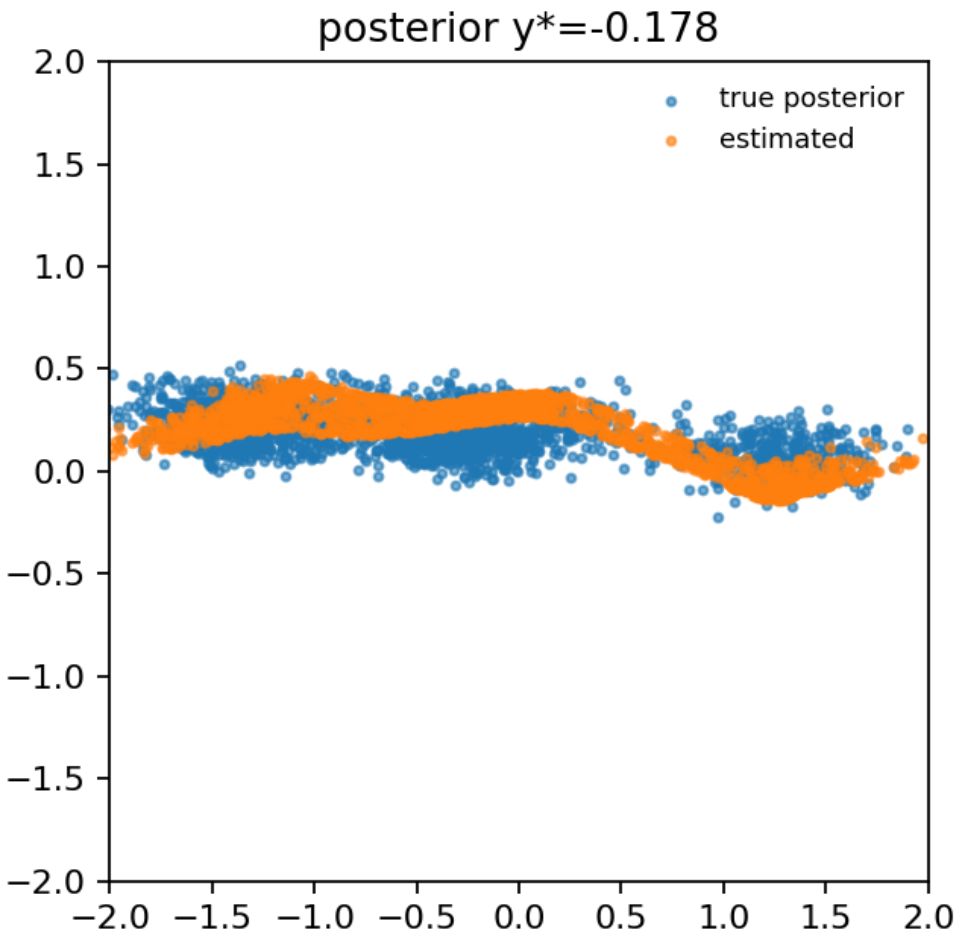
</td><td>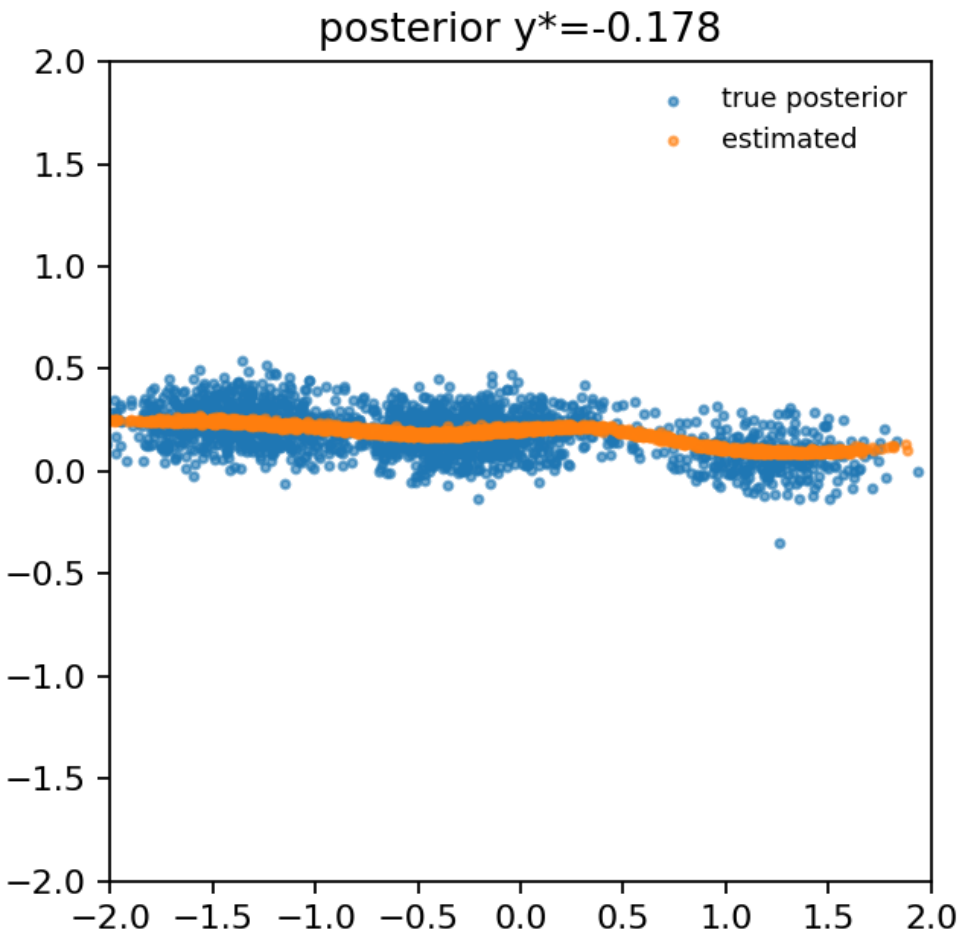
</td></tr>
</table>

Table 4: Samples of posterior distribution for two observations for paired WAE and cVAE (with analytic KL weight in ELBO). Conditional cVAE outputs collapse to 1D.